%% file: main.tex
\documentclass[11pt]{article}
\RequirePackage[colorlinks,citecolor=blue,urlcolor=blue,linkcolor=blue]{hyperref}
\usepackage{url}
\usepackage{booktabs}
\usepackage{svg}
\usepackage{tcolorbox}
\usepackage{subcaption}
\usepackage{url,amscd,paralist,bbm,wrapfig}
\usepackage{bm}
\usepackage{algorithm}
\usepackage{algorithmic} 
\usepackage{todonotes}
\usepackage{amsmath}
\usepackage{amsthm}
\usepackage{amsopn}
\usepackage{amsfonts}
\usepackage{cleveref}
\usepackage{natbib}

\title{Feature, Alignment, and Supervision in Category Learning: \\
A Comparative Approach with Children and Neural Networks}

\author{Fanxiao Wani Qiu\thanks{Department of Psychology, University of Southern California (email: \texttt{fanxiaoq@usc.edu})}
\and 
Oscar Leong\thanks{Department of Statistics and Data Science, University of California, Los Angeles (email: \texttt{oleong@stat.ucla.edu})}
}
\date{}

\begin{document}

\maketitle

\begin{abstract}
Understanding how humans and machines learn from sparse data is central to cognitive science and machine learning. Using a species-fair design, we compare children and convolutional neural networks (CNNs) in a few-shot semi-supervised category learning task. Both learners are exposed to novel object categories under identical conditions. Learners receive mixtures of labeled and unlabeled exemplars while we vary supervision (1/3/6 labels), target feature (size, shape, pattern), and perceptual alignment (high/low). We find that children generalize rapidly from minimal labels but show strong feature-specific biases and sensitivity to alignment. CNNs show a different interaction profile: added supervision improves performance, but both alignment and feature structure moderate the impact additional supervision has on learning. These results show that human–model comparisons must be drawn under the right conditions, emphasizing interactions among supervision, feature structure, and alignment rather than overall accuracy.

\end{abstract}

\maketitle

\input{sections/intro}
\input{sections/methods}

\input{sections/results}
\input{sections/discussion}

\bibliographystyle{plain}

\bibliography{main}

\end{document}

%% file: sections/intro.tex
\section{Introduction}

An impressive feat of human learning is our ability to learn from sparse data. Infants are already capable of one-shot learning, in which they learn the meaning of a new word through a single exposure and retain knowledge of the word even a week later \citep{carey1978acquiring}. In contrast, machine learning algorithms need hundreds to thousands of data points to achieve the same level of learning. This discrepancy raises a foundational question: what enables humans to learn so efficiently from limited data? Decades of developmental research suggest that efficiency is supported by our strong built-in inductive biases, including sensitivity to perceptual salience \citep{diesendruck2003specific, landau1988importance}, the whole object assumption \citep{markman1984children}, the use of social cues such as joint attention \citep{tomasello2005understanding}, and sensitivity to ostensive signaling \citep{csibra2011natural}. In the perceptual domain, certain features, most notably shape (relative to texture, size, or color), are privileged by both children and adults when extending category labels, facilitating rapid and systematic category learning early in development. Although one-shot word learning provides a striking illustration of how quickly humans learn, most real-world learning occurs in settings where only a small subset of categories are explicitly labeled, with many more encountered without labels. Thus, understanding how inductive biases guide learning in semi-supervised regimes is critical for explaining how children generalize from limited instruction in naturalistic environments. 


Modern deep neural networks provide a natural point of comparison, as they have achieved striking success in perceptual learning and can be trained under controlled mixtures of labeled and unlabeled data. While many of these earlier successes occurred in the supervised regime with a large amount of labeled data \citep{krizhevsky2012imagenet, he2016deep}, self-supervised learning algorithms based on methods such as contrastive learning \citep{chen2020simple} have showcased impressive performance when small amounts of labeled data in combination with a large amount of unlabeled data. These methods succeed not only because they use more data, but because their training objectives and architectures impose specific inductive biases that shape the structure of the learned representations. These advances have naturally raised the question of whether the learning strategies employed by modern neural networks resemble those used by human learners, and if so, in what ways they align or diverge. This question has motivated a growing body of work comparing the structure of representations \citep{peterson2018evaluating, muttenthaler2025aligning}, inductive biases \citep{geirhos2018imagenet, hermann2020origins, goyal2022inductive, feinman2018learning, tartaglini2022developmentally, tartaglini2023deep}, and generalization behavior \citep{ito2022compositional} of humans and neural networks across a range of perceptual and conceptual tasks.

Despite increasing interest in comparing deep neural networks with human learners, much of the existing literature relies on cross-study comparisons or loosely matched tasks, making it difficult to draw principled conclusions about the sources of success or failure in each system. Firestone \cite{firestone2020performance} emphasized the importance of what he terms ``species-fair" comparisons between humans and machines, with comparisons that (1) place human-like constraints on machines, (2) place machine-like constraints on humans, and (3) align tasks in a species-appropriate manner. In the present work, we implement such species-fair comparisons in category learning by examining human and machine learning in settings where the category-relevant dimensions, difficulty of examples, and level of supervision are explicitly controlled. Most prior work implementing species-fair comparisons has focused primarily on adult participants (e.g., \cite{lake2015human, kim2018not, rajalingham2018large}). While informative, this focus leaves an open gap: if inductive biases are central to human data efficiency, then examining learning early in development, before extensive experience and formal schooling, provides a potentially more direct window into the mechanisms that support efficient generalization in humans. 

\paragraph{Our approach: a semi-supervised and contrastive learning framework}

Semi-supervised learning provides a particularly informative setting for comparing human and machine learning because it closely approximates the conditions under which humans typically acquire categories. In naturalistic settings, children rarely receive labels for every novel object they encounter. Instead, they learn from a small number of informative labeled exemplars embedded within their daily lives. Recent work suggests that even infants can effectively learn categories when labeled input is sparse and embedded within a larger set of unlabeled exemplars \citep{latourrette2022sparse}. At the same time, prior work in both cognitive science and machine learning suggests that learning in semi-supervised settings is highly sensitive to task structure, with unlabeled data sometimes facilitating and sometimes hindering learning, depending on distribution and alignment (e.g., \cite{broker2022unsupervised}). Thus, success in semi-supervised learning may be less dependent on the quantity of labeled data and more on how learners infer category structure from limited labeled exemplars. 

Crucially, the learning signal available in such settings is often relational rather than categorical: learners are exposed to limited labels alongside repeated opportunities to compare whether pairs of stimuli belong to the same underlying category. In machine learning, a closely related (and empirically successful) paradigm is contrastive learning \citep{chen2020simple}, which aims to learn representations by enforcing consistency between related inputs while distinguishing unrelated ones. Evidence from cross-situational word learning \citep{smith2008infants, yu2007rapid, suanda2014cross} indicates that learners can accumulate information across noisy situations, using co-occurrence structure to resolve ambiguity even when labels are sparse. Hence a contrastive objective could act as a natural computational proxy for this type of weak supervision. Consistent with this, contrastive learning has been used to train multimodal models on egocentric child data in ways that capture meaningful word–referent mappings (e.g., \cite{vong2024grounded}), suggesting it is not only conceptually appropriate but also empirically promising for bridging child and model learning. By comparing children and modern contrastive-learning architectures side by side, we can therefore assess whether their performance reflects similar strategies for integrating labeled and unlabeled information, or whether children’s efficiency comes from  different inductive biases that constrain hypothesis space from the onset.

\paragraph{Category difficulty and structural alignment} To understand the similarities and differences between children and machines in learning categories of varying levels of difficulty, we adopt Structural Mapping Theory \citep{gentner1983structure} as a framework. A central claim of Structural Mapping Theory is that learning through comparison depends on establishing maximal structural alignment between two representations, such that corresponding elements and relations can be placed into one-to-one correspondence. Visual comparison has been shown to facilitate the discovery of conceptual and relational structure by making such correspondences salient \citep{christie2010hypotheses}. Within this framework, alignment refers to the degree to which two representations support systematic one-to-one mapping of their parts and relations. Prior work shows that such differences in alignability affect children’s ability to extract relational structure and form categories: children are faster and more accurate when comparing highly aligned pairs than poorly aligned pairs (e.g., \cite{zheng2022spatial}), and they are more likely to learn novel object names when exemplars share high structural similarity and clear correspondence \citep{gentner2007comparison}. We examine whether alignment affects both children and CNNs' performance in a semi-supervised learning setting.

\section{Current Study}

The present study directly compares children and modern contrastive-learning architectures under matched semi-supervised learning conditions to assess their category learning ability and to elucidate their sensitivity to properties such as feature structure, alignment, and level of supervision. Critically, our goal is not to determine whether children or machines perform better overall, but to identify the factors that differentially support success in each learner and how these factors potentially interact. We do so by varying (1) the feature on which categories are defined (size, shape, or pattern, as in \cite{landau1988importance}), (2) the amount of supervision available (1/6, 3/6, or 6/6 labeled exemplars), and (3) the degree of alignment among exemplars (high vs. low). To our knowledge, the present work is the first to compare child and CNNs in a controlled semi-supervised learning regime.






%% file: sections/methods.tex
\section{Methods}

\subsection{Stimuli} 

\begin{figure}[t]
  \centering
  \includegraphics[width=0.75\columnwidth]{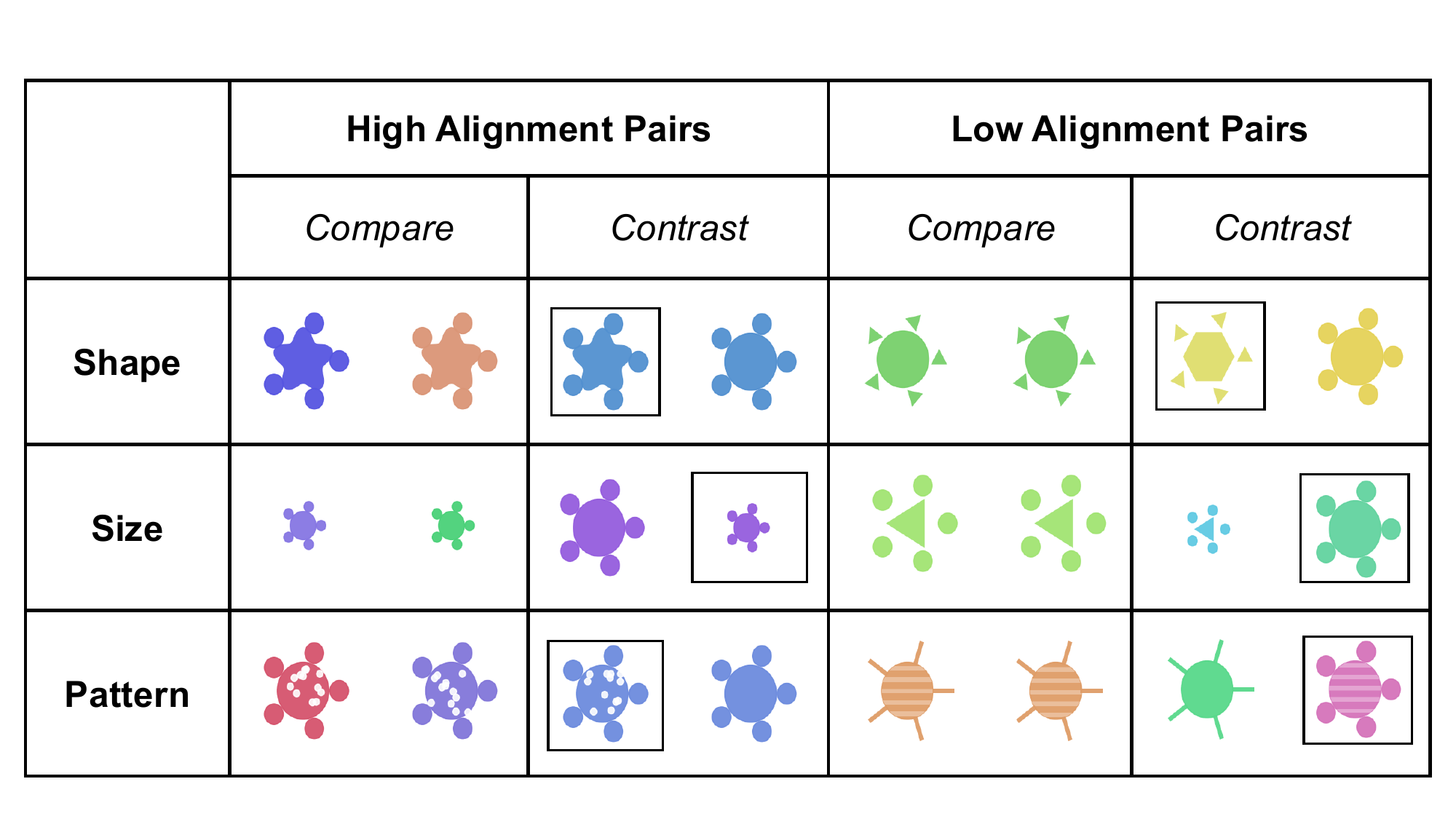}
  \caption{Examples of stimuli pairs used for each feature category (shape, size, pattern) and alignment (high, low).}
  \label{fig:stimuli}
\end{figure}

In our stimulus set, we systematically varied the feature structure and the degree of alignment. See Figure~\ref{fig:stimuli} for examples of stimuli along with how they differ in the two alignment conditions. The dimension that defined category membership is either shape (center shape or appendage shape), size (large or small), or pattern (dotted, striped, or plain). Additionally, categories were either high-alignment or low-alignment. In the high-alignment condition, paired exemplars in the contrast trials shared the same perceptual features and differed only along the target feature dimension (shape, size, or surface pattern), affording clear correspondence between the two models and making the category-relevant commonality easy to identify. Paired exemplars in compare trials shared the same target feature (e.g., dotted) but varied in color, making it easier to isolate the category-relevant dimension in the input. In the low-alignment condition, exemplars in the contrast trials differed in dimensions in addition to the target feature, reducing perceptual correspondence and making it harder to track which dimension was invariant across category members. For example, for categories defined by size, the stimuli also varied in center shape and color. In the low-alignment compare trials, positive instances of paired exemplars shared both the target feature (e.g., striped) and color, obscuring the target feature and making it harder to discern which attribute is the diagnostic one. The alignment manipulation also systematically varies the isolability of the category-relevant dimension in the input, which should affect both children and CNNs by making it easier or harder to extract the invariant structure across exemplars, even if the underlying learning mechanisms differ.

\subsection{Human participants}

Power analyses were conducted using the \texttt{simr} package in R based on pilot data from 13 participants (6 categories each) fit with the model \texttt{Percentcorrect $\sim$ Trait + Alignment + Sup\_Trials + (1|ID)}. Simulations extended the number of participant clusters while holding the item structure constant. All predictors reached at least 90\% power by $N = 21$. We decided on an $N = 24$ for counterbalancing reasons. 

The final sample consisted of 24 5- to 7-year-olds (M = 82.58 months, 50\% female). Participants were recruited from various regions of the United States. Per parental report, 54.2\% were White, 33.3\% Asian, 8.3\% African American, 4.2\% biracial; 4.2\% of the children were Latinx. Children were tested via Zoom and received an electronic gift card for participating. 

\paragraph{Materials and Design}

We employed a within-subjects, repeated measures design. Each participant completed six tasks, one for each category (Modi, Toma, Zorg, Hux, Adet, Bem). Task order (order of the six categories), number of supervised trials (1/6, 3/6, 6/6), compare or contrast (whether the pair presented is pair-same or pair-different) were counterbalanced. Gender was balanced.  

\paragraph{Procedure}


Each trial followed a fixed structure. Children first saw an introductory slide (``In this round, I will be teaching you what [category name, e.g., modis] are.") followed by six learning trials shown for 6 seconds each. Depending on counterbalancing, children viewed either pair-same or pair-different slides, each presented for exactly 6 seconds before moving onto the next pair.

For the supervised pair-same trials, E said, ``Look! These are both [modis]". For the supervised pair-different slides, E said, ``Look! Only this (as a square encompassed the target object) is a modi." For unsupervised trials, E said, ``Look at these!" 

Following the 6 learning trials, the children were presented with the classification task, during which they viewed 12 new stimuli (6 within the category and 6 outside the category) and were asked, ``Can you tell me which of these are [modis]?" 

\subsection{Machine Learning}

\paragraph{Model and training details} We employ a Siamese neural network \citep{chicco2021siamese}  that takes a single paired stimulus image and splits it into left and right halves, which are then processed by a shared visual encoder. Concretely, each half-image is passed through an 18-layer CNN with residual connections (ResNet-18 \citep{he2016deep}), producing a 512-dimensional feature vector for each half. A binary classification head maps each feature vector to a logit indicating whether that half contains the target object, and a small projection Multi-Layer Perceptron (MLP) maps the same feature vector to a latent vector used for representation learning. This shared-encoder structure ensures that the left and right crops are represented in a common feature space while still allowing separate object judgments for each side.

\begin{figure}[t]
  \centering
  \includegraphics[width=0.75\columnwidth]{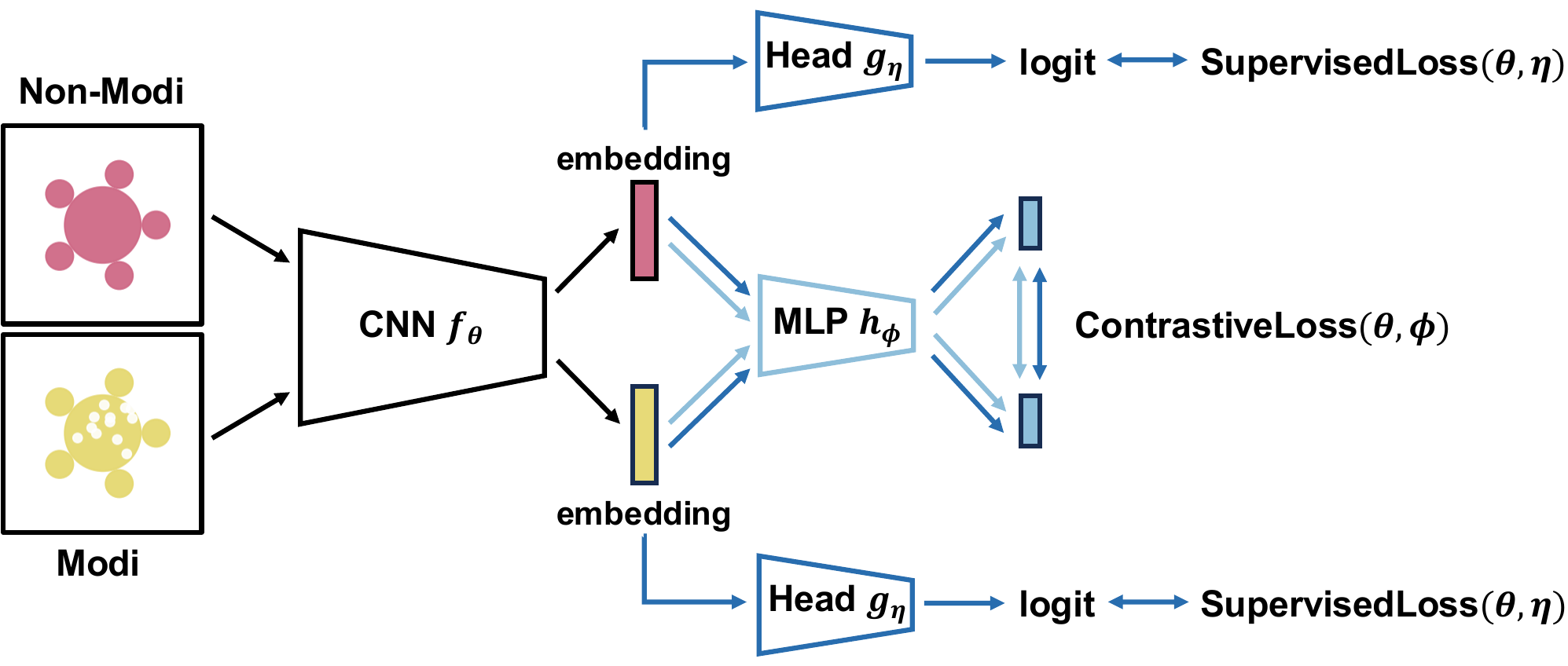}
  \caption{A pair of images is processed independently by a shared CNN $f_{\theta}$, producing embeddings for each image. Depending on whether the given pair of images have explicit supervision, these embeddings are used differently. In the supervised setting (dark blue line), a classification head $g_{\eta}$ maps each embedding to a logit, trained using a supervised loss on $(\theta,\eta)$ when labels are available. In the unsupervised setting (light blue line), the same embeddings are passed through a projection MLP $h_{\phi}$, and a contrastive loss on $(\theta,\phi)$ is used to encourage embeddings from same-class pairs to be similar and those from different-class pairs to be dissimilar. A contrastive loss is also used for supervised pairs to further improve the representations learned by the neural network.}
  \label{fig:neural-network-diagram}
\end{figure}

Training mixes limited supervised data with pair-structured weakly-labeled examples using a multi-objective loss. For supervised trials, the model receives labels for each half (object vs. non-object) and is trained with binary cross-entropy (BCE) per-half, plus a pair-consistency weakly supervised loss that encourages the two half-predictions to be compatible with whether the stimulus is a ``same'' or ``different'' pair. For unsupervised trials, the model is given paired images drawn from the remaining pool, along with a same/different indicator, and is trained with the same pair-consistency objective. The pairwise contrastive objective uses cosine similarity to ensure that pairs from the same class are pushed together in the feature space and different-class pairs are pushed apart. A schematic of our setup is provided in Figure~\ref{fig:neural-network-diagram}. We trained for $200$ epochs using the AdamW optimizer with learning rate $10^{-2}$ and a cosine annealing scheduler. We gathered results using these hyperparameters over $20$ different random seeds. Similar to the human participants, compare or contrast was counterbalanced across all levels of supervision (1/6, 3/6, 6/6) and random seeds. At test time, the network is given a pair of images and predicts, for each image independently, whether it belongs to the target category (Modi vs. non-Modi). We compute accuracy at the image level (i.e., each image is scored separately), and the paired format is used only to match the child procedure and to support the model’s pair-based training objectives.

%% file: sections/results.tex
\section{Results}

\begin{figure*}[ht]
  \centering
  \begin{subfigure}[t]{0.49\textwidth}
    \centering
    \includegraphics[width=\linewidth]{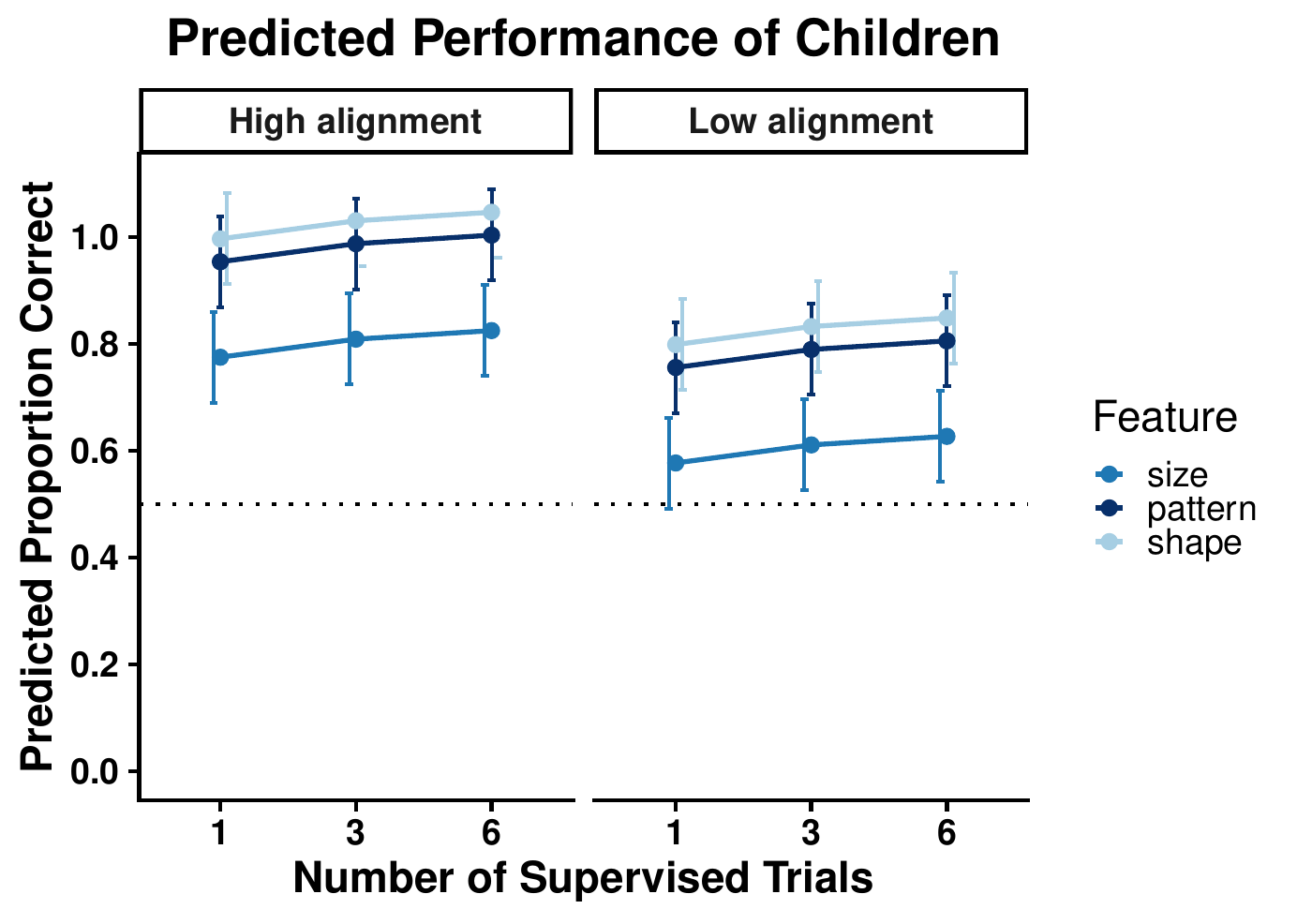}
  \end{subfigure}\hfill
  \begin{subfigure}[t]{0.49\textwidth}
    \centering
    \includegraphics[width=\linewidth]{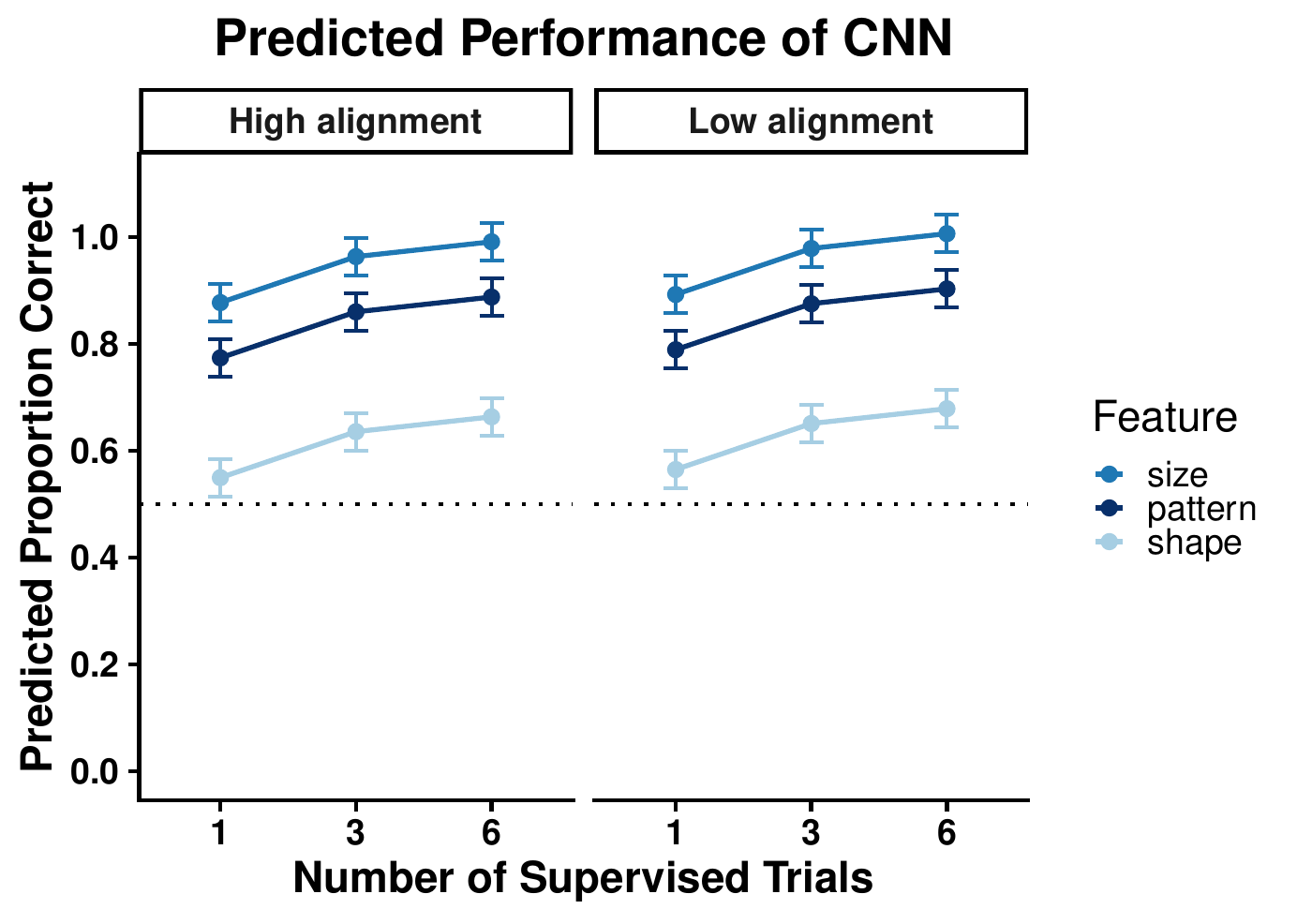}
  \end{subfigure}
   \caption{Predicted proportion correct as a function of the number of supervised trials, feature type, and alignment condition for child (left) and CNN (right), with error bars for 95\% CI. Lines show model predictions and points indicate condition means. Dotted line indicates chance performance.} 
  \label{fig:child-cnn-comparison}
\end{figure*}

\subsection{Human Participants}

A linear mixed-effects model was fit to percent correct (number of correctly identified category members out of 6) with Alignment (factor: high, low), Feature (factor: shape, size, pattern), and number of Supervised Trials (factor: 1, 3, 6) as fixed effects and random intercepts for participants (see Figure~\ref{fig:child-cnn-comparison}). We found that mean accuracy was high ($b = .775$, $p < .001$) for the intercept (high-alignment, feature size, and 1/6 supervised trials). A significant main effect of Alignment was observed. Notably, accuracy was lower for low-alignment items than for high-alignment items, ($b = -.198$, $SE = .034$, $p < .001$) suggesting that when object pairs had higher perceptual similarity, children were 19.8\% more accurate in their judgments compared to when object pairs had lower perceptual similarity. 

The defining feature also significantly affected children's performance. Relative to size, accuracy was significantly higher for shape (+22.2 percentage points, $b = .222$, $SE = .042$, $p < .001$) and for pattern (+17.9 percentage points, $b = .179$, $SE = .042$, $p < .001$).

Neither half supervision ($b = .034$, $SE = 0.042$, $p = .425$), nor full supervision ($b = .050$, $SE = 0.042$, $p = .240$), resulted in reliable improvements relative to the 1/6 trials, indicating no robust learning or practice effect once Alignment and Feature were controlled. Increasing the number of labeled exemplars yielded diminishing returns for children’s learning when controlling for diagnostic feature and alignment, with additional supervision insufficient to overcome children's difficulties with learning in size-based categories. None of the counterbalanced factors (Task order, order of compare or contrast trials, order of supervised learning trials) predicted children's responses ($ps > .14$).

\begin{table*}[t]
\centering
\scriptsize
\setlength{\tabcolsep}{1.75pt}
\renewcommand{\arraystretch}{0.9}
\begin{tabular}{c|cc|cc|cc|cc|cc|cc}
\toprule
& \multicolumn{6}{c}{High Alignment} & \multicolumn{6}{c}{Low Alignment} \\
Feature
& Human(1) & CNN(1) & Human & CNN(3) & Human(6) & CNN(6)
& Human(1) & CNN(1) & Human(3) & CNN(3) & Human(6) & CNN(6) \\
\midrule
Size
& 0.69 & 0.91 & 0.88 & 0.90 & 1.00 & 0.90
& 0.56 & 1.00 & 0.62 & 1.00 & 0.48 & 1.00 \\
Shape
& 1.00 & 0.53 & 0.96 & 0.69 & 0.96 & 0.80
& 0.79 & 0.60 & 0.94 & 0.53 & 0.90 & 0.60 \\
Pattern
& 1.00 & 0.68 & 0.94 & 0.87 & 1.00 & 0.90
& 0.81 & 0.72 & 0.73 & 0.98 & 0.82 & 0.93 \\
\bottomrule
\end{tabular}
\caption{Mean proportion correct for humans and CNNs by feature, supervision level, and alignment.}
\label{tab:accuracy-human-cnn}
\end{table*}

To examine whether the effects of feature structure depended on alignment, we conducted exploratory analysis using a linear mixed-effects model predicting percent correct from Alignment, Trait, and their interaction, with random intercepts for participants. We found a significant main effect of Alignment, with lower performance in the low-alignment condition ($b = -.30$, $SE = .06$, $p <  .001$), as well as significant main effects of Trait, with both shape- ($b = .12$, $SE = .06$, $p = .041$) and pattern-defined categories ($b = .13$, $SE = .06$, $p = .035$) learned more accurately than size-defined categories in the high-alignment condition. We also observed a significant interaction with Alignment and Trait ($b = .20$, $SE = .08$, $p = .017$), suggesting that the impact of alignment differed by feature type. This interaction remained when supervision was included as a covariate. We then conducted follow-up comparisons using estimated marginal means. Findings showed that low alignment substantially impaired learning for size-defined categories ($p <  .001$) and pattern-defined categories ($p = .001$), but not for shape ($p = .092$). Children's shape bias was thus relatively robust to disruptions in perceptual alignment, whereas learning based on size and pattern was much more fragile. An emmeans analysis against chance revealed that children performed significantly above chance in all condition combinations, with the exception of the low-alignment size-based category with only one supervised trial. Although children were able to learn size-defined categories when exemplars were clearly aligned, their performance deteriorated in low alignment conditions. These findings highlight that children’s semi-supervised category learning is shaped by the interplay between feature structure and perceptual organization, rather than by either factor alone.

Since compare-contrast pairs were unbalanced in our 1/6 and 3/6 supervised trials, we also conducted exploratory analysis examining whether the proportion of comparison versus contrast labels modulated learning within each supervision level. In the 3/6 supervision condition, a higher proportion of comparison labels was associated with higher accuracy ($b = .47$, $SE = 0.20$, $p = .027$), suggesting that children benefited more from pair-same labels (e.g., ''These are both Daxes") more than pair-different labels (e.g., ''Only this is a Dax"). We observed a marginally significant effect in the opposite direction in the 1/6 condition ($b = -.139$, $SE = .076$ , $p = .074$). We discuss this finding further in the Discussion.


\subsection{Machine Learning Results}

We showcase the estimated accuracy of the CNN across $20$ random seeds (as discussed in Methods) and computed the predicted proportion correct in Figure~\ref{fig:child-cnn-comparison}. To examine whether the effects of feature structure and perceptual alignment on learning in CNNs depended on the amount of supervision, we fit a linear model predicting classification accuracy from Trait, Alignment, number of Supervised trials, and all interactions. We observed significant main effects of Trait and supervision, with higher accuracy for size- and pattern-defined categories relative to shape, and with accuracy increasing as the number of supervised trials increased (all $ps< .001$). In contrast to children, the main effect of Alignment was weak and did not reliably predict CNN performance on its own ($b = .06$, $SE = .04$, $p = .097$).

\begin{figure}[t]
  \centering
  \includegraphics[width=0.75\columnwidth]{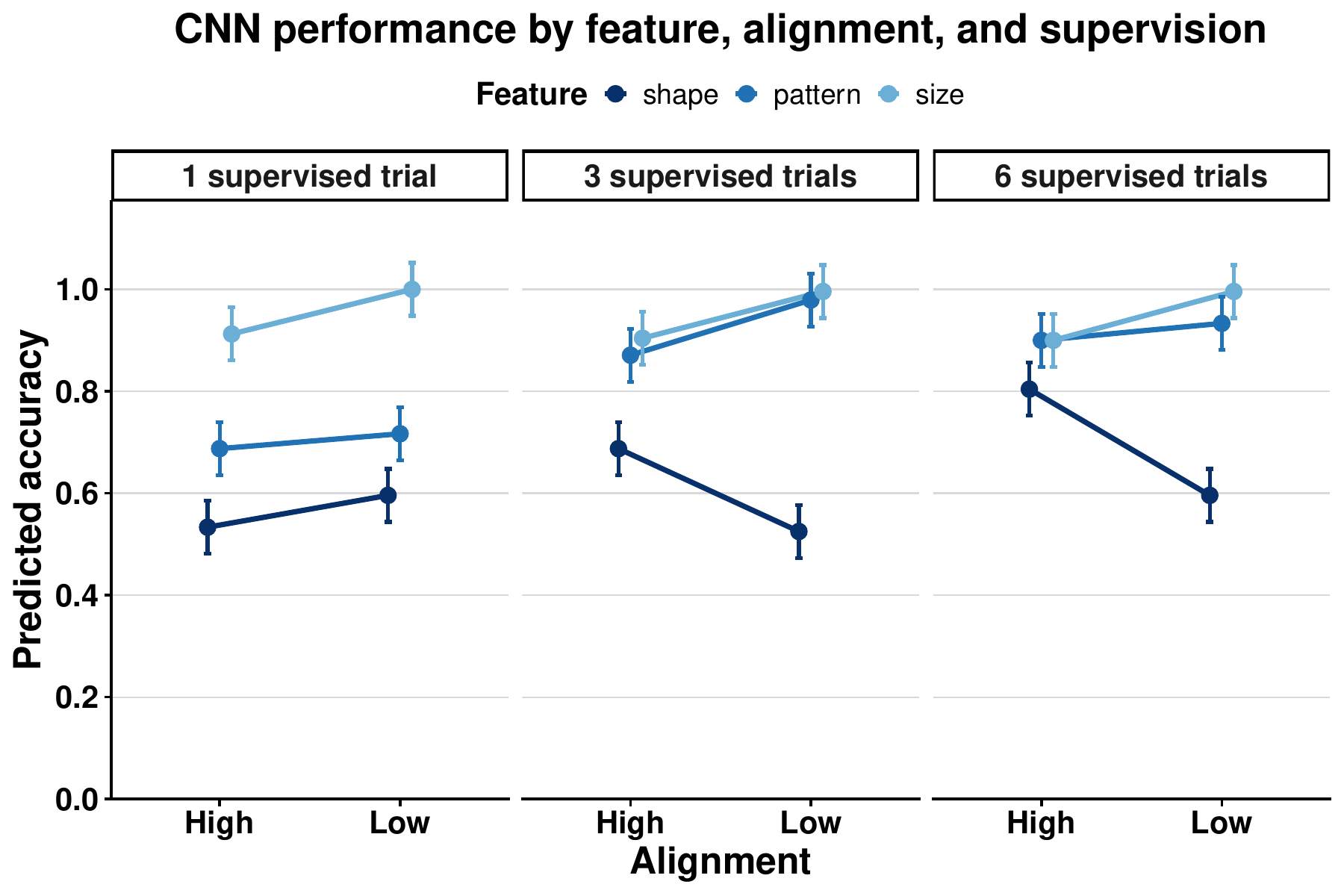}
  \caption{Predicted performance of the CNN separated by feature, alignment, and level of supervision, with error bars for 95\% CI. Lines show model predictions and points indicate condition means.}
  \label{fig:cnn-interaction}
\end{figure}

However, CNN performance was shaped by interactions among Trait, Difficulty, and Supervision (see Table~\ref{tab:accuracy-human-cnn} and Figure~\ref{fig:cnn-interaction}). Significant Trait × Supervision interactions indicated that the benefits of additional supervision were weaker for size-based categories, which were already learned well with minimal supervision (91\% accuracy for 1/6 supervised trials). Significant three way interactions suggests that the impact of supervision depended on Alignment and Trait (all $ps < .01$), demonstrating that the influence of alignment on learning depended jointly on the trait being learned and the amount of supervision available. 

Follow up analyses using marginal means revealed that alignment had a selective effect on CNN performance. For shape-defined categories, increasing supervision substantially improved performance under high alignment (from .53 at one supervised trial to .80 at six trials), whereas under low alignment, additional supervision yielded little benefit (remaining near .60 across supervision levels). In contrast, pattern-defined categories improved with supervision regardless of alignment, and size-defined categories performed near ceiling across all conditions. Thus, alignment did not have a uniform effect on CNN accuracy, but instead affected the effectiveness of supervision specifically for shape-based learning, making supervision less effective for learning when the exemplars lacked perceptual coherence. 

%% file: sections/discussion.tex
\section{Discussion}

The present study compared human learners and CNNs on a semi-supervised category learning task, systematically manipulating perceptual alignment, feature structure, and the level of supervision. Although both children and CNNs achieved above-chance performance, their learning profiles diverged in systematic ways, revealing qualitative differences in how each system acquires and generalizes category knowledge.

Children were strongly affected by perceptual alignment. Consistent with prior work, they learned categories significantly less well when object pairs exhibited low alignment, indicating reliance on perceptual similarity to learn category structure \citep{gentner1983structure, christie2010hypotheses, zheng2022spatial}. In contrast, CNN performance was largely insensitive to alignment when the feature trait was pattern and size. This could be due to the fact that CNNs primarily rely on local feature statistics that remain informative even when global spatial correspondences are disrupted, whereas humans benefit from aligned exemplars to extract relational structure. This aligns with prior evidence \citep{tartaglini2023deep} that showed in same-different relation tasks, CNNs trained from scratch can exploit data-specific shortcuts and exhibited worse generalization than vision transformers pretrained with contrastive objects. However, alignment played a significant role with respect to shape, disrupting the positive effect supervision has on learning. This could be due to the fact that in our context, CNNs did not exhibit a shape bias and the shift from high-to-low alignment further obscures the target shape feature. A deeper analysis of such issues, including comparisons with different architectures, training objectives, and the use of pretraining are important directions for future work.

Children and CNNs also exhibited opposite patterns in regards to feature structure. For children, size was the most challenging feature to learn, with performance significantly lower than for shape and pattern. Children struggled even more with learning size-based categories when alignment was low, performing at levels similar to chance. However, their shape bias remained robust even under 1/6 supervised trials and low alignment. This result converges with extensive evidence for a human shape bias \citep{diesendruck2003specific, landau1988importance}, as size is often a less stable or diagnostic cue for category membership. Future work should consider the impact of using adjectives for learning size- and pattern-based categories \citep{smith1992count}. CNNs showed the reverse pattern. Shape was the most difficult feature for the network, while performance was substantially higher for size and pattern. This finding is consistent with prior demonstrations that CNNs struggle to extract abstract shape representations \citep{kim2018not} and instead favor texture-based cues \citep{geirhos2018imagenet, hermann2020origins}. 

Another key difference emerged in the role supervision plays. Consistent with fast-mapping, children benefited little from additional labeled examples when given optimal learning conditions such as high alignment and shape as the diagnostic feature. In contrast, CNN performance improved monotonically with increased supervision, showing robust gains with three and six labeled examples relative to a single example. 

Exploratory analyses suggested that for children, the type of supervision mattered even when the amount of supervision did not. With three labeled trials, a higher proportion of comparison trials was associated with higher accuracy, whereas with only one labeled trial we observed a marginal effect in the opposite direction. This pattern indicates that the usefulness of comparison and contrast labels depends on how much category-relevant evidence learners have already accumulated. Consistent with prior work showing that comparison and contrast provide different kinds of information \citep{ankowski2013comparison}, contrast labels may be more informative when evidence is extremely sparse by marking category boundaries, whereas comparison labels become more beneficial once learners have multiple exemplars, directing attention to shared structure \citep{namy2010differing}. Future work should systematically explore the differing effects of compare and contrast under various conditions.

\paragraph{Implications for human–model comparisons}

More broadly, our results bridge findings from recent neural network accounts of inductive bias acquisition and developmental findings. \cite{feinman2018learning} show that simple neural networks can acquire a shape bias with only three to four examples, and that the emergence of this bias predicts improvements in learning efficiency, mirroring developmental trajectories in children. The present work reveals a further difference: although both children and neural networks can acquire biases that privilege certain dimensions, they differ in how those biases interact with other factors crucial to successful learning, such as structural alignment and supervision. Structural alignment interacted with feature structure differently for children and CNNs: it was crucial to children's success for size-based categories and gated the positive effect of supervision on CNN's learning of shape-based categories. This divergence suggests that while inductive bias acquisition may be achievable in current architectures, human-like use of those biases requires additional constraints on how data is integrated and compared.

These results have broader implications for how computational models are evaluated as accounts of human cognition. It is increasingly common to assess models based on whether they achieve human-level accuracy or above-chance performance on benchmark tasks. Our findings suggest that such summaries can be misleading because apparent similarities or differences hinge on which conditions are compared. In our experiments, model-human correspondence depended on interactions among supervision, target feature, and perceptual alignment. Thus, the most informative comparisons are not whether a model ``matches humans'' on average, but whether it reproduces the same pattern of tradeoffs and interaction effects across principled manipulations, which reveals when two learners succeed for the same reasons versus different ones.


%% file: main.bib
@article{carey1978acquiring,
  title={Acquiring a single new word},
  author={Carey, Susan and Bartlett, Elsa},
  journal={Papers and Reports on
Child Language Development},
  volume={15},
  pages={17--29},
  year={1978},
  publisher={ERIC}  
}

@article{diesendruck2003specific,
  title={How specific is the shape bias?},
  author={Diesendruck, Gil and Bloom, Paul},
  journal={Child development},
  volume={74},
  number={1},
  pages={168--178},
  year={2003},
  publisher={Wiley Online Library}
}

@article{gentner1983structure,
  title={Structure-mapping: A theoretical framework for analogy},
  author={Gentner, Dedre},
  journal={Cognitive science},
  volume={7},
  number={2},
  pages={155--170},
  year={1983},
  publisher={Elsevier}
}

@article{smith1992count,
  title={Count nouns, adjectives, and perceptual properties in children's novel word interpretations.},
  author={Smith, Linda B and Jones, Susan S and Landau, Barbara},
  journal={Developmental Psychology},
  volume={28},
  number={2},
  pages={273},
  year={1992},
  publisher={American Psychological Association}
}

@article{zheng2022spatial,
  title={Spatial alignment facilitates visual comparison in children},
  author={Zheng, Yinyuan and Matlen, Bryan and Gentner, Dedre},
  journal={Cognitive Science},
  volume={46},
  number={8},
  pages={e13182},
  year={2022},
  publisher={Wiley Online Library}
}

@article{namy2010differing,
  title={The differing roles of comparison and contrast in children’s categorization},
  author={Namy, Laura L and Clepper, Lauren E},
  journal={Journal of Experimental Child Psychology},
  volume={107},
  number={3},
  pages={291--305},
  year={2010},
  publisher={Elsevier}
}

@article{christie2010hypotheses,
  title={Where hypotheses come from: Learning new relations by structural alignment},
  author={Christie, Stella and Gentner, Dedre},
  journal={Journal of Cognition and Development},
  volume={11},
  number={3},
  pages={356--373},
  year={2010},
  publisher={Taylor \& Francis}
}

@article{gentner2007comparison,
  title={Comparison facilitates children's learning of names for parts},
  author={Gentner, Dedre and Loewenstein, Jeffrey and Hung, Barbara},
  journal={Journal of Cognition and Development},
  volume={8},
  number={3},
  pages={285--307},
  year={2007},
  publisher={Taylor \& Francis}
}

@article{landau1988importance,
  title={The importance of shape in early lexical learning},
  author={Landau, Barbara and Smith, Linda B and Jones, Susan S},
  journal={Cognitive development},
  volume={3},
  number={3},
  pages={299--321},
  year={1988},
  publisher={Elsevier}
}

@article{markman1984children,
  title={Children's sensitivity to constraints on word meaning: Taxonomic versus thematic relations},
  author={Markman, Ellen M and Hutchinson, Jean E},
  journal={Cognitive psychology},
  volume={16},
  number={1},
  pages={1--27},
  year={1984},
  publisher={Elsevier}
}

@article{tomasello2005understanding,
  title={Understanding and sharing intentions: The origins of cultural cognition},
  author={Tomasello, Michael and Carpenter, Malinda and Call, Josep and Behne, Tanya and Moll, Henrike},
  journal={Behavioral and brain sciences},
  volume={28},
  number={5},
  pages={675--691},
  year={2005},
  publisher={Cambridge University Press}
}

@article{csibra2011natural,
  title={Natural pedagogy as evolutionary adaptation},
  author={Csibra, Gergely and Gergely, Gy{\"o}rgy},
  journal={Philosophical Transactions of the Royal Society B: Biological Sciences},
  volume={366},
  number={1567},
  pages={1149--1157},
  year={2011},
  publisher={The Royal Society}
}

@article{firestone2020performance,
  title={Performance vs. competence in human--machine comparisons},
  author={Firestone, Chaz},
  journal={Proceedings of the National Academy of Sciences},
  volume={117},
  number={43},
  pages={26562--26571},
  year={2020},
  publisher={National Academy of Sciences}
}

@article{lake2015human,
  title={Human-level concept learning through probabilistic program induction},
  author={Lake, Brenden M and Salakhutdinov, Ruslan and Tenenbaum, Joshua B},
  journal={Science},
  volume={350},
  number={6266},
  pages={1332--1338},
  year={2015},
  publisher={American Association for the Advancement of Science}
}

@article{kim2018not,
  title={Not-So-CLEVR: learning same--different relations strains feedforward neural networks},
  author={Kim, Junkyung and Ricci, Matthew and Serre, Thomas},
  journal={Interface focus},
  volume={8},
  number={4},
  pages={20180011},
  year={2018},
  publisher={The Royal Society}
}

@article{rajalingham2018large,
  title={Large-scale, high-resolution comparison of the core visual object recognition behavior of humans, monkeys, and state-of-the-art deep artificial neural networks},
  author={Rajalingham, Rishi and Issa, Elias B and Bashivan, Pouya and Kar, Kohitij and Schmidt, Kailyn and DiCarlo, James J},
  journal={Journal of Neuroscience},
  volume={38},
  number={33},
  pages={7255--7269},
  year={2018},
  publisher={Society for Neuroscience}
}

@article{latourrette2022sparse,
  title={Sparse labels, no problems: Infant categorization under challenging conditions},
  author={LaTourrette, Alexander and Waxman, Sandra R},
  journal={Child development},
  volume={93},
  number={6},
  pages={1903--1911},
  year={2022},
  publisher={Wiley Online Library}
}

@article{broker2022unsupervised,
  title={When unsupervised training benefits category learning},
  author={Br{\"o}ker, Franziska and Love, Bradley C and Dayan, Peter},
  journal={Cognition},
  volume={221},
  pages={104984},
  year={2022},
  publisher={Elsevier}
}

@article{krizhevsky2012imagenet,
  title={Imagenet classification with deep convolutional neural networks},
  author={Krizhevsky, Alex and Sutskever, Ilya and Hinton, Geoffrey E},
  journal={Advances in neural information processing systems},
  volume={25},
  year={2012}
}

@inproceedings{he2016deep,
  title={Deep residual learning for image recognition},
  author={He, Kaiming and Zhang, Xiangyu and Ren, Shaoqing and Sun, Jian},
  booktitle={Proceedings of the IEEE conference on computer vision and pattern recognition},
  pages={770--778},
  year={2016}
}

@inproceedings{chen2020simple,
  title={A simple framework for contrastive learning of visual representations},
  author={Chen, Ting and Kornblith, Simon and Norouzi, Mohammad and Hinton, Geoffrey},
  booktitle={International conference on machine learning},
  pages={1597--1607},
  year={2020},
  organization={PmLR}
}

@article{ito2022compositional,
  title={Compositional generalization through abstract representations in human and artificial neural networks},
  author={Ito, Takuya and Klinger, Tim and Schultz, Doug and Murray, John and Cole, Michael and Rigotti, Mattia},
  journal={Advances in neural information processing systems},
  volume={35},
  pages={32225--32239},
  year={2022}
}

@article{goyal2022inductive,
  title={Inductive biases for deep learning of higher-level cognition},
  author={Goyal, Anirudh and Bengio, Yoshua},
  journal={Proceedings of the Royal Society A},
  volume={478},
  number={2266},
  pages={20210068},
  year={2022},
  publisher={The Royal Society}
}

@inproceedings{feinman2018learning,
  title={Learning Inductive Biases with Simple Neural Networks},
  author={Feinman, Reuben and Lake, Brenden M},
  booktitle={Proceedings of the Annual Meeting of the Cognitive Science Society},
  volume={40},
  year={2018}
}

@article{muttenthaler2025aligning,
  title={Aligning machine and human visual representations across abstraction levels},
  author={Muttenthaler, Lukas and Greff, Klaus and Born, Frieda and Spitzer, Bernhard and Kornblith, Simon and Mozer, Michael C and M{\"u}ller, Klaus-Robert and Unterthiner, Thomas and Lampinen, Andrew K},
  journal={Nature},
  volume={647},
  number={8089},
  pages={349--355},
  year={2025},
  publisher={Nature Publishing Group UK London}
}

@article{peterson2018evaluating,
  title={Evaluating (and improving) the correspondence between deep neural networks and human representations},
  author={Peterson, Joshua C and Abbott, Joshua T and Griffiths, Thomas L},
  journal={Cognitive science},
  volume={42},
  number={8},
  pages={2648--2669},
  year={2018},
  publisher={Wiley Online Library}
}

@inproceedings{geirhos2018imagenet,
  title={ImageNet-trained CNNs are biased towards texture; increasing shape bias improves accuracy and robustness},
  author={Geirhos, Robert and Rubisch, Patricia and Michaelis, Claudio and Bethge, Matthias and Wichmann, Felix A and Brendel, Wieland},
  booktitle={International conference on learning representations},
  year={2018}
}

@article{chicco2021siamese,
  title={Siamese neural networks: An overview},
  author={Chicco, Davide},
  journal={Artificial neural networks},
  pages={73--94},
  year={2021},
  publisher={Springer}
}

@article{vong2024grounded,
  title={Grounded language acquisition through the eyes and ears of a single child},
  author={Vong, Wai Keen and Wang, Wentao and Orhan, A Emin and Lake, Brenden M},
  journal={Science},
  volume={383},
  number={6682},
  pages={504--511},
  year={2024},
  publisher={American Association for the Advancement of Science}
}

@article{smith2008infants,
  title={Infants rapidly learn word-referent mappings via cross-situational statistics},
  author={Smith, Linda and Yu, Chen},
  journal={Cognition},
  volume={106},
  number={3},
  pages={1558--1568},
  year={2008},
  publisher={Elsevier}
}

@article{yu2007rapid,
  title={Rapid word learning under uncertainty via cross-situational statistics},
  author={Yu, Chen and Smith, Linda B},
  journal={Psychological science},
  volume={18},
  number={5},
  pages={414--420},
  year={2007},
  publisher={SAGE Publications Sage CA: Los Angeles, CA}
}

@article{suanda2014cross,
  title={Cross-situational statistical word learning in young children},
  author={Suanda, Sumarga H and Mugwanya, Nassali and Namy, Laura L},
  journal={Journal of experimental child psychology},
  volume={126},
  pages={395--411},
  year={2014},
  publisher={Elsevier}
}

@article{ankowski2013comparison,
  title={Comparison versus contrast: Task specifics affect category acquisition},
  author={Ankowski, Amber A and Vlach, Haley A and Sandhofer, Catherine M},
  journal={Infant and Child Development},
  volume={22},
  number={1},
  pages={1--23},
  year={2013},
  publisher={Wiley Online Library}
}

@article{hermann2020origins,
  title={The origins and prevalence of texture bias in convolutional neural networks},
  author={Hermann, Katherine and Chen, Ting and Kornblith, Simon},
  journal={Advances in neural information processing systems},
  volume={33},
  pages={19000--19015},
  year={2020}
}

@article{tartaglini2023deep,
  title={Deep neural networks can learn generalizable same-different visual relations},
  author={Tartaglini, Alexa R and Feucht, Sheridan and Lepori, Michael A and Vong, Wai Keen and Lovering, Charles and Lake, Brenden M and Pavlick, Ellie},
  journal={arXiv preprint arXiv:2310.09612},
  year={2023}
}

@inproceedings{tartaglini2022developmentally,
  title={A Developmentally-Inspired Examination of Shape versus Texture Bias in Machines},
  author={Tartaglini, Alexa R and Vong, Wai Keen and Lake, Brenden},
  booktitle={Proceedings of the Annual Meeting of the Cognitive Science Society},
  volume={44},
  number={44},
  year={2022}
}
